\documentclass[10pt,twocolumn,letterpaper]{article}

\usepackage{cvpr}
\usepackage{times}
\usepackage{epsfig}
\usepackage{graphicx}
\usepackage{amsmath}
\usepackage{amssymb}

\usepackage{multirow}

\usepackage{algorithm}
\usepackage{algorithmic}
\usepackage{balance}

\usepackage[dvipsnames]{xcolor}
\newif\ifcomments
\usepackage[pagebackref=true,breaklinks=true,letterpaper=true,colorlinks,bookmarks=false]{hyperref}

\cvprfinalcopy % *** Uncomment this line for the final submission

\def\cvprPaperID{5703} % *** Enter the CVPR Paper ID here

\ifcvprfinal\fi
\begin{document}

%%%%%%%%% TITLE
%\title{CAMera: Sane high-resolution image saliency for deep classifiers}
\title{CAMERAS: Enhanced Resolution And Sanity preserving Class Activation Mapping for image saliency}
\vspace{-5mm}
%%%%%%%%% AUTHORS
\author{Mohammad A. A. K. Jalwana \hspace{4mm} Naveed Akhtar  \hspace{4mm}  Mohammed Bennamoun \hspace{4mm} Ajmal Mian\\
Computer Science and Software Engineering,
The University of Western Australia.\\
\tt\small \{mohammad.jalwana@research., naveed.akhtar@, mohammed.bennamoun@, ajmal.mian@\}uwa.edu.au
\\
\tt\small Code: \url{https://github.com/VisMIL/CAMERAS}
}
\maketitle
\thispagestyle{empty}
\vspace{-2mm}
%%%%%%%%% ABSTRACT
\begin{abstract}
\vspace{-3mm}
Backpropagation image saliency aims at explaining model predictions by estimating model-centric importance of individual pixels in the input. However, class-insensitivity of the earlier layers in a network only allows saliency computation with low resolution activation maps of the deeper layers, resulting in compromised  image saliency. Remedifying this can lead to sanity failures. We propose CAMERAS, a technique to compute high-fidelity backpropagation saliency maps without requiring any external priors and preserving the map sanity. Our method  systematically performs multi-scale accumulation and fusion of the activation maps and   backpropagated gradients to compute precise saliency maps. From accurate image saliency to articulation of relative importance of input features for different models, and precise discrimination between model perception of visually similar objects, our high-resolution mapping offers multiple novel insights into the black-box deep visual models, which are presented in the paper. 
We also demonstrate the utility of our saliency maps in adversarial setup by drastically reducing the norm of attack signals by focusing them on the precise regions identified by our maps. Our method also inspires new evaluation metrics and a sanity check for this developing research direction.

%Attribution problem is concerned with identification of  input parts that govern model decisions. Numerous techniques have been proposed in recent years, however they often  compromise visual quality of explanation maps with model fidelity. In this paper, we introduce an algorithm that bridges this artificial trade-off and enable novel and interesting insights about decision of visual model. Our technique enables visual model to analyze a given image at variety of scales and iteratively accumulates internal layers activations and gradients. The resultant explanation mask captures the salient regions that appear in close vicinity to human concept, pass sanity checks, clarifies model association in presence of similar objects, highlight relationship among different architectures and enable a natural defense against maligned inputs. 

\end{abstract}

\vspace{-5mm}
%%%%%%%%% BODY TEXT
\section{Introduction}
\label{sec:Intro}
% \begin{itemize}
%     \item dont need any instrumentation of the netwok
% \end{itemize}
\vspace{-1.5mm}
Deep visual models are fast surpassing human-level performance for various vision tasks, including image classification \cite{krizhevsky2012imagenet}, \cite{simonyan2014very}, object detection~\cite{redmon2017yolo9000}, \cite{ren2015faster}, and semantic segmentation \cite{long2015fully}, \cite{chen2018encoder}.
However, they hardly offer any explanation of their decisions, and are rightfully considered black-boxes.
This is problematic for their practical deployment, especially in high-risk emerging applications where transparency is vital, e.g.~in healthcare, self-driving vehicles and smart surveillance~\cite{rebuffi2020there}. 
The problem is exacerbated by the push of `right to explanation' by algorithmic regulatory authorities and their objection to black-box models in safety-critical applications~\cite{adadi2018peeking}. 
%Deep learning has enabled computer vision algorithms to achieve or even surpass human level performance in various tasks, including image classification \cite{krizhevsky2012imagenet, simonyan2014very}, object detection~\cite{redmon2017yolo9000, ren2015faster}, semantic segmentation \cite{long2015fully, chen2018encoder} and image captioning~\cite{wu2017image}.
%However, despite their impressive accuracy, these models do not offer any explanation of their decision and are rightfully considered black-boxes.
%Explanation can play vital role to ensure the transparency and keep the models free of any bias. For example, the  detailed manual analysis of pneumonia patient admission system revealed counter-intuitive patterns leveraged by models for higher accuracy. 
%The opaqueness of models has thus pushed algorithmic regulators to promote `right to explanation` and object to black-box models deployment in critical applications like disease prognostics and smart surveillance \textcolor{red}{[Refs.]}. 

Addressing this issue for deep visual models, techniques are emerging to offer input-agnostic~\cite{jalwana2020attack} and input-specific~\cite{fong2019understanding}, \cite{rebuffi2020there}, \cite{selvaraju2017grad}  explanation of model predictions. This work subscribes to the latter, where the ultimate goal is to identify the contribution of each pixel in an input to the output prediction. 
The popular techniques to achieve this adopt one of two strategies.
%a bicephalous approach. 
%The existing literature mainly takes a bicephalous approach to achieve this objective.
The first, systematically modifies the input image pixels (i.e.~image regions) and analyzes the effects of those perturbations on the output predictions~\cite{fong2019understanding}, \cite{fong2017interpretable}, \cite{petsiuk2018rise}, \cite{zeiler2014visualizing}. The underlying \textit{search} nature of this perturbation-based formulation offers high-fidelity model-centric importance attribution to the input pixels, albeit at a high computational cost. Hence, tractability is achieved under heuristics or external priors over the computed importance maps. This is undesired because the eventual maps may be influenced by these external factors, which compromises the model-fidelity of the maps.

The second  strategy relies on the activation maps of the  internal layers and gradients of the models. Commonly known as backpropagation saliency methods~\cite{selvaraju2017grad}, \cite{rebuffi2020there},  \cite{simonyan2014deep}, \cite{springenberg2014striving}, \cite{zeiler2014visualizing}, \cite{zhang2018top} approaches adopting this strategy are computationally  efficient, thereby offering the possibility of avoiding unnecessary heuristics or priors. However, for the visual neural models, the layers closer to the input are class-insensitive~\cite{rebuffi2020there}. This limits the ammunition of backpropagation saliency methods to the deeper layers of the networks, where the size of activation maps is very small, e.g.~$10^{-3}\times$ of the input size. Projecting the saliency computed with those maps onto the original image grid results in intrinsically low-resolution image saliency. On the other hand, using heuristics or priors to sharpen those projections  inadvertently compromise the sanity of the  maps~\cite{adebayo2018sanity}. Not to mention, employing activation signals of multiple internal layers for  resolution enhancement takes us back to a combinatorial search problem of choosing the best layers, under a pre-specified heuristic.              

\begin{figure*}[t!]
    \centering
    \includegraphics[width=0.87\textwidth]{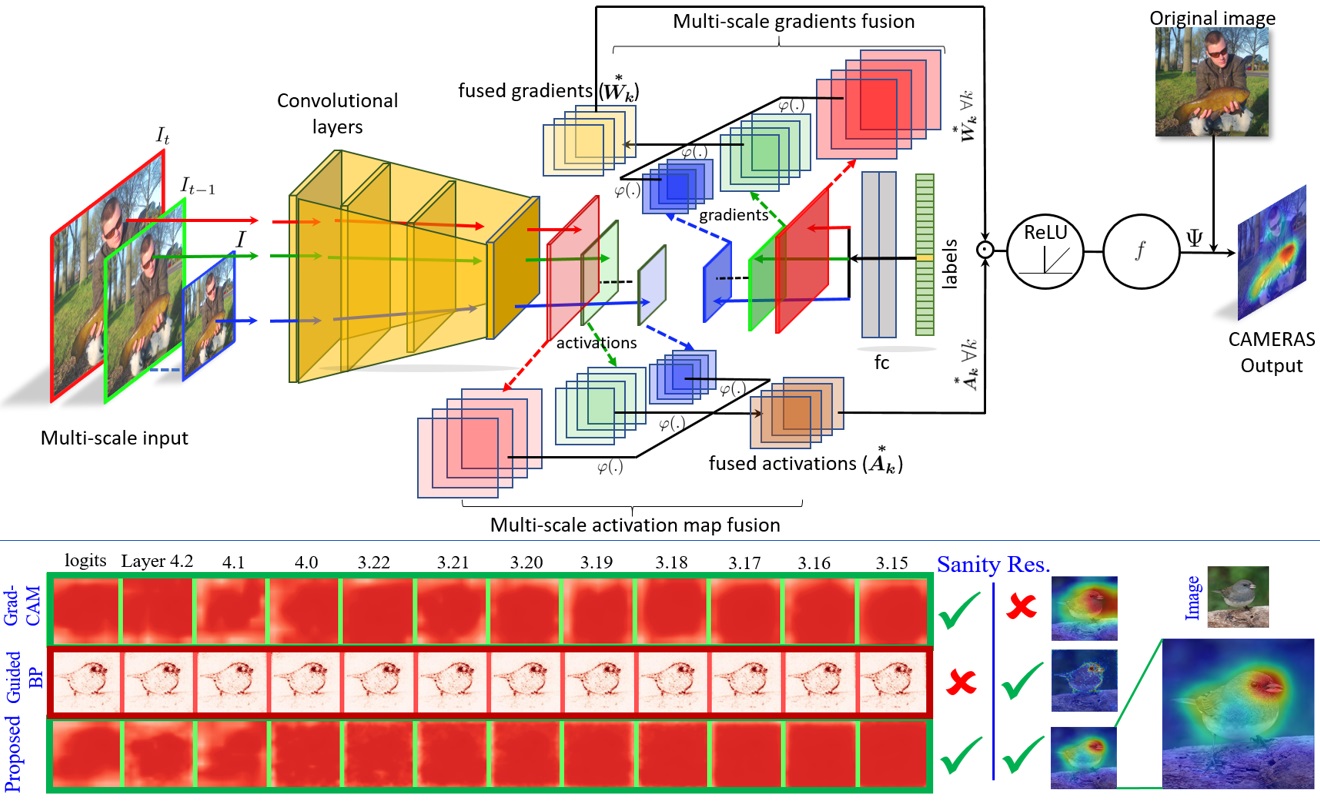}
    \caption{(\textbf{Top}) CAMERAS meticulously fuses activation maps and backpropagated gradients of a layer for multi-scale copies of an input. After passing the resulting saliency map through ReLU and normalising it ($f$), the map is embedded on the original image. (\textbf{Bottom}) By avoiding influence of external factors, CAMERAS easily passes the sanity checks for image saliency. Shown are the results of cascading randomization~\cite{adebayo2018sanity} on ResNet. Progressive randomising of layer weights randomises the output right from the logits layer which identifies preservation of sanity. Thus, the CAMERAS maps do not sacrifice their sanity for high-resolution, achieving the best of both worlds.}
    \label{fig:main}
    \vspace{-3mm}
\end{figure*}

Addressing the above issues, we introduce \textbf{CAMERAS} - an \textbf{E}nhanced \textbf{R}esolution \textbf{A}nd \textbf{S}anity  preserving \textbf{C}lass \textbf{A}ctivation \textbf{M}apping for backpropagation image saliency. The proposed technique (Fig.~\ref{fig:main}-top) systematically accumulates and fuses multi-scale activation maps and backpropagated gradients of a model to construct precise saliency maps. By avoiding the influence of any  external factor, e.g.~heuristics, priors, thresholds, the saliency maps of CAMERAS easily pass the sanity checks for image saliency (Fig.~\ref{fig:main}-bottom). Moreover, the technique allows saliency estimation with a single network layer, not requiring any layer search for map resolution enhancement. Contributions of the paper are summarised below:
\vspace{-1mm}
\begin{itemize}
\setlength\itemsep{-0.1em}
    \item We propose CAMERAS for precise backpropagation image saliency while preserving the sanity. Our method outperforms the  state-of-the-art saliency methods by a large margin, achieving up to $27.5\%$ error reduction for the popular pointing game metric~\cite{zhang2018top}. 
    \item Exploring the newly found precision saliency mapping with CAMERAS, we visualise  differences in the semantic understanding of different architectures that govern their performance. We also highlight model-centric discrimination of input features for visually similar objects in never-before-seen details. 
    \item Considering the equivalent treatment of deep models as differentiable programs by the fast-developing parallel  field of adversarial learning, we enhance the widely considered strongest adversarial attack  PGD~\cite{madry2017towards} with our saliency technique - drastically improving the efficacy of the  attack.
    \item The ability of precise saliency computation allowed by CAMERAS calls for new quantitative metrics and sanity checks. We contribute two new evaluation metrics and a sanity check to advance this research direction. 
\end{itemize}

\vspace{-3mm}
\section{Related work}
\vspace{-1mm}
The literature has seen multiple techniques that perturb input pixels and measure its effects on the model outputs to identify salient regions in input images~\cite{fong2017interpretable}, \cite{fong2019understanding}, \cite{petsiuk2018rise}. Perturbing every possible combination of pixels has an exponential complexity. Therefore, such techniques often rely on fixed subsets of pixel combinations for  tractability. Moreover, high non-linearity of deep models further restricts the saliency map to be reliable under  fixed devised perturbation subsets. 
RISE \cite{petsiuk2018rise} and Occlusion \cite{zeiler2014visualizing} generate attribution maps by weighing perturbation masks corresponding to the changes in the output scores. 
Other techniques, such as Meaningful perturbations \cite{fong2017interpretable}, Extremal perturbations \cite{fong2019understanding}, Real-time saliency \cite{dabkowski2017real} and \cite{ribeiro2016should}  cast the problem into an optimization objective. Though effective, these methods face a common issue of allowing a channel for external influence on the resulting maps in the form of e.g.~heuristics, external constraints, priors or threshold etc.   

Backpropagation saliency methods \cite{simonyan2014deep}, \cite{selvaraju2017grad}, \cite{zeiler2014visualizing} aim at extracting information from within the model to identify model-centric salient regions in an input image. Relying on layer activations and model gradients, these methods are also known to be computationally more effective \cite{zintgraf2017visualizing}, \cite{kapishnikov2019xrai}. 
% The other class of methods, relevant to our work, is backpropagation based saliency techniques that are computationally efficient as they require a fixed number of network traversal.
Simoyan et al.~\cite{simonyan2014deep} first used model gradients as a possible explanation of output predictions. Different adaptations have since been proposed to mitigate  the inherent noise sensitivity of model gradients. Guided back prop~\cite{springenberg2014striving} and DeConvNet \cite{zeiler2014visualizing} alter the backpropagation rules of model ReLU layers, while SmoothGrad \cite{smilkov2017smoothgrad} computes the average gradients over samples in the close vicinity of the original one. Similarly, DeepLIFT \cite{shrikumar2017learning}, LRP \cite{bach2015pixel} and Excitation Backprop \cite{zhang2018top} recast the backpropagation rules such that the sum of attribution signal becomes unity. Sundararajan et al.~\cite{sundararajan2017axiomatic} interpolated multiple attribution maps to reduce the signal noise. There are also instances of exploring saliency computation using various layers of deep models by merging the layer activation maps with the  gradient information. Such methods include CAM \cite{zhou2016learning}, its generalized adaption GradCAM \cite{selvaraju2017grad}, linear approximation \cite{kindermans2016investigating} and NormGrad \cite{rebuffi2020there}. For such methods, it is found that the layers closer to output generate better saliency maps because those layers are more sensitive to the high level class features.  

Beyond the visual quality of saliency maps, a few works have also critically explored the reliability of these maps for different methods~\cite{mahendran2016salient}, \cite{kindermans2019reliability}.  Adebayo et al.~\cite{adebayo2018sanity} first introduced sanity checks for image saliency methods, highlighting that visual appeal of the maps alone can be misleading. They evaluated sensitivity of the results  of popular techniques to model parameters. Surprisingly, `model-centric saliency maps' computed by multiple methods were found insensitive to the model - failing the sanity check. The techniques avoiding external influences on the map, e.g.~Grad-CAM~\cite{selvaraju2017grad} easily passed the test. This finding also resonated with other subsequent sanity checks~\cite{rebuffi2020there}. 

%As shown in Figure \textcolor{red}{[cite]} that our technique successfuly passes this sanity check and unlike GradCAM does not compromise the finer details in the attribution map. However, beside high resolution, Guided Backpropagation \textcolor{red}{[Ref]} remains insensitive to  parameters randomization. 

Evaluation of image saliency methods is a challenging problem because deep model representation is not always aligned with the human visual  system~\cite{tsipras2018robustness}. Hence, indirect evaluation of image saliency is often done by analysing its weak localization performance. 
%Since each image may have multiple objects so such performance measure also considers class sensitivity.
For instance, Pointing game score~\cite{zhang2018top} is a commonly used metric for quantitative evaluation of image saliency results \cite{fong2019understanding}, \cite{rebuffi2020there}. It measures the correlation between the maximal point in a saliency map with the semantic labels of the pixel. Its later adaptation  \cite{fong2017interpretable} measures the overlap in the bounding boxes from saliency maps and the ground truth. Petsiuk et al.~\cite{petsiuk2018rise} proposed insertion-deletion metrics to measure the impact of perturbations over image patches in the order of importance to quantify saliency accuracy. Nevertheless, these metrics are meant to be weak indicators of saliency maps due to the imprecise nature of the maps computed by the earlier methods, which is no longer the case for CAMERAS. 

\section{Proposed Approach}
\label{sec:PA}
\vspace{-1mm}
Before discussing the details of our technique, we first provide a closer look at the broader paradigm of backpropagation saliency computation. We use  Grad-CAM~\cite{selvaraju2017grad} - a popular technique - as a test case to motivate the proposed method. The text below highlights only the relevant aspects of the test case for intuition.
%We refer to~\cite{selvaraju2017grad} for details.      

% \vspace{1mm}
% \noindent {\bf Backpropagation saliency computation:} 
\vspace{-1mm}
\subsection{Saliency computation with backpropagation}
\label{sec:BPSC}
\vspace{-1mm}
Let $\boldsymbol{I} \in \mathbb R^{c\times h \times w}$ be an input image with `$c$' channels. A deep visual classifier $\mathcal K (\boldsymbol{I})$ maps $\boldsymbol{I}$ to a prediction vector $\boldsymbol{y}^{\ell} \in \mathbb R^L$, where `$L$' is the total number of classes. Here, `$\ell$' indicates the predicted label of $\boldsymbol{I}$ under the premise that the $\ell^{\text{th}}$ coefficient of $\boldsymbol{y}$ has the largest value. It is well-known that a neural network is a hierarchical composition of representation layers. Rebuffi et al.~\cite{rebuffi2020there} demonstrated that among these layers, those closer to the input learn class-insensitive features. Thus, the deeper layers hold more promise for computing image saliency for a model. Grad-CAM~\cite{selvaraju2017grad} takes a pragmatic approach to single out the last convolutional layer to estimate the saliency map. 

Let us denote the $k^{\text{th}}$ activation map of the last convolutional layer of a network as $\boldsymbol{A}_k (\boldsymbol{I}) \in \mathbb R^{m \times n}$. Focusing on Grad-CAM, the technique first  computes an intermediate representation $\boldsymbol{S} (\boldsymbol{I}) \in \mathbb R^{m \times n}$, such that $\boldsymbol{S}^{(i,j)} (\boldsymbol{I}) = \sum_{k} w_{k} \boldsymbol{A}^{(i,j)}_{k} (\boldsymbol{I})$, where $w_k$ is given by Eq.~(\ref{eq:GradCAM}). Henceforth, we ignore the argument  $(\boldsymbol I)$ for clarity, unless required. 
\begin{align}
w_{k} = \frac{1}{(m+n)}\sum\limits_{i = 1}^{m} \sum\limits_{j = 1}^{n}\left(\frac{\partial \boldsymbol{y}^{\ell}}{\partial \boldsymbol{A}_{k}^{(i,j)}}\right).
\label{eq:GradCAM}
\end{align}
In the above expressions, $\boldsymbol X^{(i,j)}$ is the $(i,j)$ coefficient of $\boldsymbol X$. The computed $\boldsymbol{S}$ is later extended  to the final saliency map $\boldsymbol\Psi \in \mathbb R^{h \times w}$, as $f(\boldsymbol S): \boldsymbol S\rightarrow \boldsymbol\Psi$, where the function $f(.)$ must account for interpolating an $m\times n$ matrix for a $h\times w$ grid (along other complementary transformations). 

% \vspace{1mm}
% \noindent{\bf Sub-optimality of backpropagation methods:}
\subsection{Sub-optimality of backpropagation methods}
\label{sec:SBPM}
\vspace{-1mm}
Observing Grad-CAM (and similar methods) from the above perspective reveals two performance drain-holes in backpropagation image saliency computations.

\vspace{0.5mm}
\noindent \textit{(a) Over-simplification of the weights} $w_k$: Since $\boldsymbol{A}_k$ is an activation map, individual coefficients of this matrix should have different importance for the final prediction. Indeed, this is also reflected in the values of individual backpropagated gradients for these coefficients - computed with the expression in the parenthesis in Eq.~(\ref{eq:GradCAM}). Since the ultimate objective of image saliency is to compute importance of \textit{individual} pixels, loosing information with over-simplification of $w_k$ is not conducive. Grad-CAM takes an extreme approach of representing  $w_k$ with a scalar value. The main reason for that is, it is actually detrimental to plainly replace $w_k$ with an encoding  $\boldsymbol{W}_k \in \mathbb R^{m \times n}$ such that $\boldsymbol S = \sum_k \boldsymbol W_k \odot \boldsymbol A_k$. Here, $\odot$ is the point-wise product and $\boldsymbol{W}_k$ encodes individual backpropagated  gradients. Gradients are extremely sensitive to signal variations. Hence, even a small activation change can result in (a misleading) exaggerated weight alteration for the activation map, resulting in incorrect image saliency. Grad-CAM is able to mitigate this problem by averaging the gradients in Eq.~(\ref{eq:GradCAM}). However, this remedy comes at the cost of loosing the fine-grained information about the gradients. 

Though centered around Grad-CAM, the above discussion points to a simple, yet powerful generic notion for effective backpropagation saliency computation. That is, to better leverage the backpropagated gradients,  the differential information of the gradients (in $\boldsymbol{W}_k$) is still at our disposal to exploit. Fusing activation maps with this information promises more precise image saliency.

\vspace{0.5mm}
\noindent{\textit{(b) Large interpolated segments:}} Typically, the activation maps of the deeper convolutional layer in visual classifiers are (spatially) much smaller than the input images. For instance, in  ResNet \cite{he2016deep}, the $7\times 7$ maps of the last convolutional layer are $1024$ times smaller than the $224\times 224$ inputs. Thus, for $m \ll h$ and $n \ll w$, a saliency map $\boldsymbol\Psi$ computed with the class sensitive deeper layers must be mainly composed of interpolated segments. To contextualize, in the above  ResNet example, $99.9\%$ of the values are  \textit{generated} by the function $f(\boldsymbol S): \boldsymbol S\rightarrow \boldsymbol \Psi$ in Grad-CAM. This automatically renders $\boldsymbol\Psi$ a low-resolution map, leaving alone the issue of correctness of the importance assigned to the individual pixels in the eventual saliency map. 
%This observation is easily verifiable with the blob-like structure of the final heat maps for Grad-CAM {\color{red}(see \S~\ref{})}. 
The low resolution of saliency maps has also spawned methods to improve $f(.)$~\cite{adebayo2018sanity}, \cite{selvaraju2017grad}. However, those techniques inevitably rely on external information (including  heuristics) for the transform $\boldsymbol S\rightarrow \boldsymbol\Psi$ due to unavailability of further useful information from the model itself. This leads to sanity check failures  because the operands are no longer purely grounded in the original model.

\vspace{-1.5mm}
% \noindent{\bf Take-home note:} 
\subsection{The room for improvement}
\label{sec:THN}
\vspace{-1.5mm}
From the above discussion, it is clear that whereas useful techniques exist for backpropagation image saliency, the paradigm is yet to fully harness the backpropagated gradients and resolution enhancement of the activation maps for precise image saliency.
Both limitations are rooted in the very nature of the underlying signals. %The latter is physically restricted by the low resolution of a model's internal representation. 
Leveraging these signals from multiple layers can potentially help in partially overcoming the issues. However, this possibility is also restricted by the  class-insensitivity of the earlier network layers and combinatorial nature of the problem. Moreover, there is  evidence that multi-layer fusion can often adversely affect image saliency~\cite{rebuffi2020there}. 
Hence, a technique specifically targeting the class-sensitive last layer, while allowing  minimal loss of differential information across backpropagated gradients and improving activation map upsampling, holds significant promises for better saliency computation.

% \vspace{1mm}
% \noindent{\bf Proposed CAMeras:}

\subsection{CAMERAS}
\label{sec:PCAM}
\vspace{-1.5mm}
Building directly on the insights in \S~\ref{sec:THN}, we devise CAMERAS - an enhanced resolution and sanity preserving class activation mapping scheme. The approach is illustrated in Fig.~(\ref{fig:main}-top) and explained below in a top-down manner, keeping the flow of the above discussion.

Our method eventually computes a saliency map as:
\begin{align}
    \boldsymbol\Psi = f\left(\text{ReLU}\big(\sum_k \boldsymbol {\overset{*}{W_k}} \odot \boldsymbol {\overset{*}{A_k}} \big) \right), \forall{k},
    \label{eq:our}
\end{align}
where $\boldsymbol {\overset{*}{W_k}} \in \mathbb R^{h \times w}$ encodes the differential information of the backpropagated gradients for the $k^{\text{th}}$ activation map in a network layer, $\boldsymbol {\overset{*}{A_k}} \in \mathbb R^{h \times w}$ is an enhanced resolution encoding for the activation map itself, and $f(.)$ performs an  element-wise normalisation in the range $[0,1]$. The $\boldsymbol {\overset{*}{W_k}}$ and $\boldsymbol {\overset{*}{A_k}}$ are defined as follows
\begin{align}
   \boldsymbol{\overset{*}{W_k}} = \underset{t}{\mathbb E} \big[ \varphi_t \big( \boldsymbol W_k(\varphi_t(\boldsymbol I, \zeta_t)), \zeta_o \big)\big],\\ \nonumber \boldsymbol{\overset{*}{A_k}} = \underset{t}{\mathbb E} \big[ \varphi_t \big( \boldsymbol A_k(\varphi_t(\boldsymbol I, \zeta_t)), \zeta_o \big)\big],
\end{align}
where $\varphi_t(\boldsymbol X, \zeta_t)$ is the $t^{\text{th}}$ up-sampling applied to resize $\boldsymbol X$ to the dimensions $\zeta_t$ - provided as a tuple. We fix $\zeta_o = (h, w)$ for $\boldsymbol I \in \mathbb R^{c\times h \times w}$. This will be explained shortly. 
We compute the $(i,j)$ coefficient of $\boldsymbol{W}_k$ as $\left(\frac{\partial \boldsymbol{y}^{\ell}}{\partial \boldsymbol{A}_{k}^{(i,j)}}\right)$. The overall process of generating an image saliency map with CAMERAS is summarized as Algorithm~\ref{alg:main}.   

The algorithm computes the desired saliency map $\boldsymbol\Psi$ by an iterative multi-scale accumulation of activation maps and gradients for the $\kappa^{\text{th}}$ layer of the model. In the $t^{\text{th}}$ iteration, the input image $\boldsymbol I$ gets up-sampled to $\zeta_t$ based on the maximum desired size  $\zeta_m$ and the number of steps $N$ allowed to reach that size  (\textit{lines 4,5}). Provided that the input up-scaling does not alter the model prediction, the activation maps and backpropagated gradients to the $\kappa^{\text{th}}$ layer are also up-sampled and stored. We show this on \textit{lines 6-10} of the algorithm. Notice that, we use calligraphic symbols to distinguish 3D tensors from matrices (e.g.~$\mathcal A$ instead of $\boldsymbol A$ for activations) in the algorithm for clarity. The newly introduced symbol $\nabla \mathcal J(\kappa, \ell)$ on \textit{line 9} denotes the collective backpropagated gradients to the $\kappa^{\text{th}}$ layer $w.r.t.$~the predicted label $\ell$. Also notice, on \textit{lines 8, 9}, up-sampling of the activation maps and gradients are performed to match the original image size $\zeta_o$. This is because, the same accumulated signals are eventually transformed into the saliency map of the original image. 
We iteratively accumulate the up-sampled activation maps and gradients, and finally compute their averages on \textit{lines 12 and 13}. On \textit{line 14}, we compute the saliency map by solving Eq.~(\ref{eq:our}). Here, matrix notation is intentionally used to match the original equation.

\newcommand{\Exp}[1]{\underset{#1}{\mathbb E}}
\begin{algorithm}[t]
 \caption{CAMERAS algorithm}
 \label{alg:main} 
 \begin{algorithmic}[1]
 \renewcommand{\algorithmicrequire}{\textbf{Input:}}
 \renewcommand{\algorithmicensure}{\textbf{Output:}}
 \REQUIRE  Classifier $\mathcal K$, image $\boldsymbol{I} \in \mathbb R^{c\times h \times w}$, maximum size $\zeta_m$, steps $N$, interpolation function $\varphi(.)$, layer $\kappa$
 \ENSURE Image saliency map $\boldsymbol{\Psi} \in \mathbb R^{h \times w}$
 
 \STATE Initialize $\mathcal{A}_o$, $\mathcal{W}_o$ to $\boldsymbol{0}$ tensors, $\zeta_o = (h,w)$ and $t = t_m = 0$, $ \ell = \mathcal{K}(\boldsymbol I)$ 
\WHILE {$ t \leq  N$} 
\STATE $t \leftarrow  t+1$ 
\STATE $\zeta_t \leftarrow \zeta_{t-1} + \lfloor{\frac{\zeta_m}{N} \rfloor} (t-1)$
\STATE $\boldsymbol I_t \leftarrow \varphi_t(\boldsymbol I, \zeta_t)$
\IF {$ \mathcal{K}(\boldsymbol I_t) \rightarrow \ell$}
\STATE $t_m \leftarrow t_m + 1 $
\STATE $\mathcal{A}_t \leftarrow \mathcal{A}_{t-1} + \varphi_t \big ( \mathcal{A}(\boldsymbol I_t, \kappa), \zeta_o \big )$
\STATE $\mathcal{W}_t \leftarrow \mathcal{W}_{t-1} + \varphi_t \big ( \nabla \mathcal J(\kappa, \ell), \zeta_o \big)$
\ENDIF
\ENDWHILE
\STATE $\overset{*}{\mathcal A} \leftarrow \mathcal{A}_t / t_m$
\STATE $\overset{*}{\mathcal W} \leftarrow \mathcal{W}_t / t_m$
\STATE  $\boldsymbol\Psi = f\left(\text{ReLU}(\sum_k \boldsymbol {\overset{*}{W_k}} \odot \boldsymbol {\overset{*}{A_k}}) \right), \forall{k}$
  \STATE return 
 \end{algorithmic}
 \end{algorithm}

In Algorithm~\ref{alg:main}, CAMERAS is shown to expect four input parameters, along with the classifier and the image. We discuss the choice of $\varphi(.)$ in \S~\ref{sec:SPAS} where we eventually propose to keep this function fixed. The algorithm  optionally allows $\kappa$ for computing saliency maps using layers other than the last convolutional layer of the model. For all the experiments presented in the main paper, we keep $\kappa$ fixed to the last convolutional layer. This is due to the well-known class-sensitivity of the deeper layers of CNNs~\cite{rebuffi2020there}. Essentially, the only choice to be made is for the values of parameters $\zeta_m$ and $N$, which are related as $\zeta_m = c N \zeta_o$, where $c$ is the step size. Trading-off performance with efficiency, the choice of these parameters is mainly governed by the available computational resources. For larger $\zeta_m$ and $N$ values, performance of CAMERAS roughly improves monotonically - generally saturating at $\zeta_m \approx (1K, 1K)$ for $\zeta_o = (224, 224)$ for the popular ImageNet models. For this $\zeta_m$ range, the performance is largely insensitive for $N \in [5, 10]$. \textcolor{black}{We give further analysis of parameter values in the supplementary material}. In the presented experiments, we empirically choose $\zeta_m = (1K, 1K)$ and $N = 7$.       

 \vspace{-3mm}
% \noindent{\bf Sanity preservation and strength}
\subsubsection{Sanity preservation and strength}
\label{sec:SPAS}
 \vspace{-1.5mm}
In CAMERAS, we do not impose any prior over the saliency map, nor we use any heuristic to guide its computation process. The technique also preserves model fidelity by requiring no structural (or any other) alteration to the original model. It mainly relies on primitive arithmetic operations over the model signals. These attributes also characterize those other methods that pass the popular sanity checks for backpropagation saliency~\cite{adebayo2018sanity}, \cite{rebuffi2020there}, albeit resulting in low-resolution saliency maps. In CAMERAS, the only source of any `potential' external influence on the resulting map is through the interpolation function $\varphi(.)$. We conjecture that as long as $\varphi(.)$ is a first-order function defined over the signals originating in the model itself, CAMERAS maps will always preserve their sanity because the maps would fully originate in the model. To preclude any unintentional prior over the maps, our formulation dictates the use of simpler functions as $\varphi(.)$. Hence, we choose to fix bi-linear interpolation as $\varphi(.)$.   

To analyse the reasons of extraordinary performance of CAMERAS (see \S~\ref{sec:Eva}), we provide a brief  theoretical perspective on the accumulation of multi-scale interpolated signals exploited by our method, using the results below.

% \vspace{0.5mm}
\noindent{\textit{Lemma 3.1}:} \textit{For $\overset{*}{\boldsymbol X} \in \mathbb R^{h \times w}$  and its interpolated approximation $\widetilde{\boldsymbol X} = \varphi(\boldsymbol X)$ s.t.~$ \boldsymbol X \in \mathbb R^{m \times n}$ and $\overset{*}{\boldsymbol X} \neq {\widetilde{\boldsymbol X}}$,  $||\overset{*}{\boldsymbol X} - \widetilde{\boldsymbol X}|| = f\left( (h-m),(w-n) \right)$ for $m < h$ and $n < w$, where $\varphi(.)$ denotes bi-linear interpolation and $f(.)$ is a monotonic function over its arguments.}

% \vspace{0.5mm}
\noindent{\textit{Lemma 3.2}:} \textit{For $\widetilde{\boldsymbol X}_z = \varphi(\boldsymbol X_z)$, where $\boldsymbol X_z \in \mathbb R^{m \times n}$ and $\widetilde{\boldsymbol X}_p = \varphi(\boldsymbol X_p)$, where $\boldsymbol X_p \in \mathbb R^{p \times q}$ s.t.~$p < m$ and $n < q$, $|| \overset{*}{\boldsymbol X} - \widetilde{\boldsymbol X}_z|| -  || \overset{*}{\boldsymbol X}- \widetilde{\boldsymbol X}_p|| \leq 0$.} 

\vspace{0.5mm}
\noindent{\textbf{Corollary:}} $ \underset{z}{\mathbb E} [ || \overset{*}{\boldsymbol X} - \widetilde{\boldsymbol X}_z|| ] \leq || \overset{*}{\boldsymbol X}- \widetilde{\boldsymbol X}_p||, \forall \widetilde{\boldsymbol X}_z \cup \widetilde{\boldsymbol X}_p$.

\vspace{0.5mm}
\noindent The \textit{lemma 3.1}  states that bi-linear interpolation tends to be more accurate when the difference in the dimensions of the source signal and the target grid is smaller.  The \textit{lemma~3.2} can also be easily verified  as $\widetilde{\boldsymbol X}_z$ is at least as accurate a projection of $\overset{*}{\boldsymbol X}$ as $\widetilde{\boldsymbol X}_p$, according to \textit{lemma~3.1}. This necessarily makes its error equal to or smaller than the error of  $\widetilde{\boldsymbol X}_p$. In the light of \textit{lemma~3.2}, the  corollary affirms that the expected error of a set of interpolated signals is upper-bounded by the error of the least accurate signal in the set.  

The above analysis highlights an important aspect. The CAMERAS results will necessarily be as accurate as operating our algorithm on  the original input size, and will improve monotonically thereof with the up-sampled inputs. This is because the activations and gradients of the up-sampled inputs would map more accurately on the original image grid, as per the above results.   
This is significant because it allows CAMERAS to use the differential information in the backpropagated gradient maps while accounting for the noise-sensitivity of the gradients by averaging out these signals across multiple scales. This is the key strength of the proposed technique.   

\begin{figure*}[t!]
    \centering
    \includegraphics[width=0.9\textwidth]{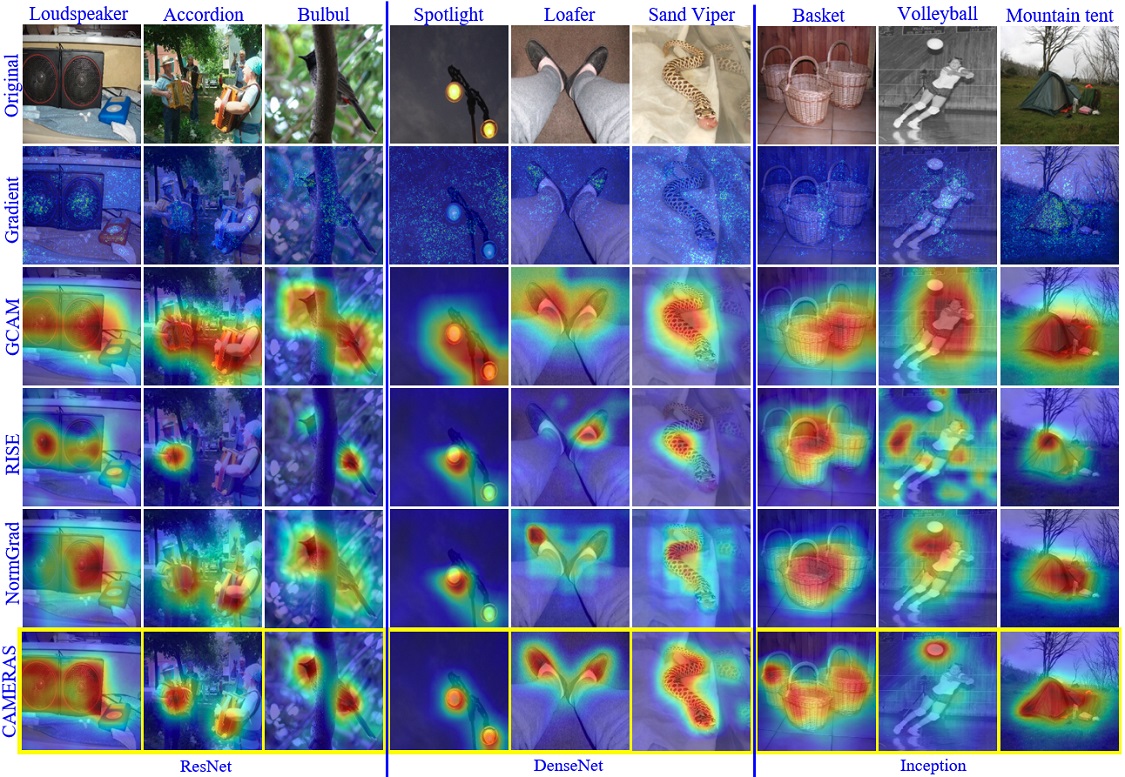}
    \caption{Visual comparison of saliency maps with the state-of-the-art methods. Representative maps are shown for random class images for {ResNet-50}, {DenseNet-121} and {Inception-v3}. Class labels are provided at the top. Results of proposed CAMERAS are in yellow box.} 
    \label{fig:comparisonResults}
    \vspace{-4mm}
\end{figure*}

\vspace{-3mm}
\section{Evaluation}
\label{sec:Eva}
\vspace{-2mm}
We perform a thorough qualitative and quantitative evaluation of CAMERAS on large-scale models and compare the performance with the state-of-the-art methods.  

\vspace{-2mm}
\subsection{Qualitative results}
\label{sec:QR}
\vspace{-2mm}
In Fig.~\ref{fig:comparisonResults}, we qualitatively compare the results of our technique with the existing saliency methods for ImageNet models of ResNet, DenseNet and Inception. 
Representative maps of randomly chosen images are provided.
\textcolor{black}{See the supplementary material for more visualisations}.
The high quality of CAMERAS maps is apparent in the figure. A quick inspection reveals that our technique maintains its performance across a variety of scenarios, including clear objects (Loudspeaker, Mountain tent), occluded objects (Bulbul), and multiples instances of objects (Accordion, Basket). Observing carefully, CAMERAS maps provide precise maps even for small and relatively complex geometric shapes, e.g.~Volleyball, Spotlight, Loafer,  Hognose snake. Interestingly, our method is able to attach appropriate importance even to the reflection of the Hognose snake. Adoption of saliency maps to complex geometric shapes is a direct consequence of enabling precise saliency mapping while sealing-off any external influence on the maps. CAMERAS is able to maintain its characteristic precision across different models and images. These are highly promising results for explainability of modern deep visual classifiers. 

%Figure \ref{fig:comparisonResults} highlights that our technique compactly captures the salient regions even when there are multiple instances such as in `Accordion', `Spot light' and `loafer'. Interestingly, for 'Hognose snake', all techniques unianimously agree on the skin patterns to be of prime importance. However, our algorithm as well as `GradNorm' (partially) highlight the reflection pattern in the mirror to be highlighted as well. Further, our algorithm is robust to oclussions where other techniques are unable to account for the background region. This can be observed for the case of `Bulbul'.

% \begin{enumerate}
% \item  Main experiment to compare our with others
% \item Application - Association
% \item Application - Networks relation
% \item Attack and Defense - (Low norm attack, more transferable)
% \end{enumerate}

\vspace{-2mm}
\subsection{Quantitative results}
\label{sec:QuanR}
\vspace{-2mm}
A quantum leap in performance with CAMERAS is also observed  in our quantitative results. Saliency maps are typically evaluated by measuring their correlation with the semantic annotations of  the image. Pointing game \cite{zhang2018top} is a popular metric for that purpose, which considers the computed saliency for every object class in the image. If the maximal point in the saliency map is contained within the object, it is considered a hit; otherwise, a miss. The performance is measured as the percentage of successful hits. We refer to \cite{zhang2018top} for more details on the metric.   
Table \ref{tablePointingGame} benchmarks the  performance of CAMERAS for  pointing game on $4,952$ images of PASCAL VOC test set \cite{pascal-voc-2007}, and $\sim50$K images of COCO 2014 validation set~\cite{lin2014microsoft}. Our technique consistently shows superior performance, achieving up to $27.5\%$ error reduction. The gain is higher for ResNet as compared to VGG due the better  performance of the original ResNet that permits better  saliency.   
%We argue that default spatial resolution of VGG-16 is double than the ResNet models so our multiresolution technique assists ResNet higher than VGG-16. 

\begin{table}
\centering
\resizebox{\columnwidth}{!}{%
\begin{tabular}{ p{2.0cm} p{1.5cm} p{1.5cm} p{1.5cm} p{1.5cm}}
 \hline
       & \multicolumn{2}{c}{\textit{\underline{VOC07 Test (All/Diff)}}} & \multicolumn{2}{c}{\textit{\underline{COCO14 Val (All/Diff)}}}  \\
\textbf{Method} & \textbf{\textit{VGG16}} & \textbf{\textit{ResNet50}} & \textbf{\textit{VGG16}} & \textbf{\textit{ResNet50}} \\
 \hline
Center~\cite{fong2019understanding}         & 69.6/42.4 & 69.6/42.4 & 27.8/19.5 & 27.8/19.5 \\
Gradient~\cite{simonyan2014deep}            & 76.3/56.9 & 72.3/56.8 & 37.7/31.4 & 35.0/29.4 \\
DeConv.~\cite{zeiler2014visualizing}        & 67.5/44.2 & 68.6/44.7 & 30.7/23.0 & 30.0/21.9 \\
Guid~\cite{springenberg2014striving}        & 75.9/53.0 & 77.2/59.4 & 39.1/31.4 & 42.1/35.3 \\
MWP~\cite{zhang2018top}                     & 77.1/56.6 & 84.4/70.8 & 39.8/32.8 & 49.6/43.9 \\
cMWP~\cite{zhang2018top}                    & 79.9/66.5 & 90.7/82.1 & 49.7/44.3 & 58.5/53.6 \\
$\mathrm{RISE^{\ast}}$~\cite{petsiuk2018rise}             & 86.9/75.1 & 86.4/78.8 & 50.8/45.3 & 54.7/50.0 \\
GradCAM~\cite{selvaraju2017grad}            & 86.6/74.0 & 90.4/82.3 & 54.2/49.0 & 57.3/52.3 \\
$\mathrm{Extremal^{\ast}}$~\cite{fong2019understanding}   & \textbf{88.0}/76.1 & 88.9/78.7 & 51.5/45.9 & 56.5/51.5 \\
NormGrad~\cite{rebuffi2020there}            & 81.9/64.8 & 84.6/72.2 & -  &  - \\
\textbf{CAMERAS}                                        & 86.2/\textbf{76.2} & \textbf{94.2}/\textbf{88.8} & \textbf{55.4/50.7}&
\textbf{69.9}/\textbf{66.4}  \\
\hline
\end{tabular}
}
\caption{Mean accuracy on pointing game over the full data (\textit{All}) splits and subset of difficult images (\textit{Diff}), as specified in \cite{zhang2018top}. The results of other schemes are generated with TorchRay package~\cite{fong2019understanding}, and `*' denotes an average over 3 runs for improved performance.}
\label{tablePointingGame}
\end{table}

% \begin{table}
% \centering
% \resizebox{\columnwidth}{!}{%
% \begin{tabular}{ p{1.2cm} p{1.0cm} p{1.0cm} p{1.0cm} p{1.0cm} | p{1.0cm} p{1.0cm} p{1.0cm} p{1.0cm} } \hline 
%         & \multicolumn{4}{c}{\textit{\underline{Positive map density ($\rho^+_{\text{map}} \uparrow$ )}}} & \multicolumn{4}{c}{\textit{\underline{Negative map density ($\rho^-_{\text{map}} \downarrow$})}} \\
%  \textbf{Model} &  \textbf{\textit{NoMask}} &  \textbf{\textit{NGrad}} & \textbf{\textit{GCAM}} &\textbf{\textit{Ours}} &
%  \textbf{\textit{NoMask}} &  \textbf{\textit{NGrad}} & \textbf{\textit{GCAM}}  & \textbf{\textit{Ours}  } \\ \hline
%  ResNet      & 0.87 &  1.67    & 2.33 & \textbf{3.20} &  -  &  0.96      & 0.86  & \textbf{0.81} \\
%  DenseNet    & 0.84 &  1.76    & 2.35  & \textbf{3.23} &  - &  1.02      & 0.94  & \textbf{0.83} \\
%  Inception   & 0.94 &  2.19    & 2.18  & \textbf{3.15} &  - &  0.95      & 1.04  & \textbf{0.93}  \\
% \end{tabular}
% }
% \label{tableScores}
% \caption{Proposed metric scores on ImageNet validation set for the saliency maps of Norm-Grad (NGrad)~\cite{rebuffi2020there},  Grad-CAM (GCAM)~\cite{selvaraju2017grad},  and our method. {\color{red}REMOVE NO MASK COL.}}
% \vspace{-3mm}
% \end{table}

\begin{table}
\centering
\resizebox{\columnwidth}{!}{%
\begin{tabular}{ p{1.5cm} | p{1.0cm} p{1.0cm} p{1.0cm} | p{1.0cm} p{1.0cm} p{1.0cm} } \hline 
        & \multicolumn{3}{c}{\textit{\underline{Positive map density ($\rho^+_{\text{map}} \uparrow$ )}}} & \multicolumn{3}{c}{\textit{\underline{Negative map density ($\rho^-_{\text{map}} \downarrow$})}} \\
 \textbf{Model}  &  \textbf{\textit{NGrad}} & \textbf{\textit{GCAM}} &\textbf{\textit{Ours}} &
 \textbf{\textit{NGrad}} & \textbf{\textit{GCAM}}  & \textbf{\textit{Ours}  } \\ \hline
 ResNet       &  1.67    & 2.33 & \textbf{3.20}  &  0.96      & 0.86  & \textbf{0.81} \\
 DenseNet     &  1.76    & 2.35  & \textbf{3.23} &  1.02      & 0.94  & \textbf{0.83} \\
 Inception    &  2.19    & 2.18  & \textbf{3.15} &  0.95      & 1.04  & \textbf{0.93}  \\
\end{tabular}
}
\caption{The proposed metric scores on ImageNet validation set for the saliency maps of Norm-Grad (NGrad)~\cite{rebuffi2020there},  Grad-CAM (GCAM)~\cite{selvaraju2017grad},  and our method.}
\label{tableScores}
%{\color{red}REMOVE NO MASK COL.}
\vspace{-4mm}
\end{table}

% \begin{table}
% \centering
% \resizebox{\columnwidth}{!}{%
% \begin{tabular}{ p{1.2cm} p{1.0cm} p{1.0cm} p{1.0cm} p{1.0cm} | p{1.0cm} p{1.0cm} p{1.0cm} p{1.0cm} } \hline 
%         & \multicolumn{4}{c}{\textit{\underline{Positive map density ($\rho^+_{\text{map}} \uparrow$ )}}} & \multicolumn{4}{c}{\textit{\underline{Negative map density ($\rho^-_{\text{map}} \downarrow$})}} \\
%  \textbf{Model} &  \textbf{\textit{NoMask}} &  \textbf{\textit{NGrad}} & \textbf{\textit{GCAM}} &\textbf{\textit{Ours}} &
%  \textbf{\textit{NoMask}} &  \textbf{\textit{NGrad}} & \textbf{\textit{GCAM}}  & \textbf{\textit{Ours}  } \\ \hline
%  ResNet      & 0.87 &  1.67    & 2.33 & \textbf{3.20} &  -  &  0.96      & 0.86  & \textbf{0.81} \\
%  DenseNet    & 0.84 &  1.76    & 2.35  & \textbf{3.23} &  - &  1.02      & 0.94  & \textbf{0.83} \\
%  Inception   & 0.94 &  2.19    & 2.18  & \textbf{3.15} &  - &  0.95      & 1.04  & \textbf{0.93}  \\
% \end{tabular}
% }
% \label{tableScores}
% \caption{The table compares proposed metrics scores for the saliency masks from GradCAM \cite{selvaraju2017grad}, NormGrad \cite{rebuffi2020there} and entire image (No Mask). The results are reported over imagenet pretrained ResNet-50, DenseNet-121 and Inception-v3.}
% \vspace{-3mm}
% \end{table}

\vspace{-2mm}
The pointing game generally disregards precision of the  saliency maps by focusing only on the maximal points. Arguably, crudeness of the saliency maps computed by the earlier methods influenced this evaluation metric. The possibility of precise saliency computation (by CAMERAS) calls for new metrics that account for the finer details of  saliency maps. We propose `positive map density' and `negative map density' as two suitable metrics, respectively defined as:
$\rho^+_{\text{map}} = P(\mathcal{K}(\boldsymbol I \odot \boldsymbol\Psi))/ \sum_i\sum_j {\boldsymbol\Psi^{(i,j)}} \times (h\times w)$, and $\rho^-_{\text{map}} = P(\mathcal{K}(\boldsymbol I \odot \boldsymbol{1}- \boldsymbol\Psi))/ \sum_i\sum_j {(\boldsymbol{1}-\boldsymbol\Psi^{(i,j)}}) \times (h\times w)$. Here, $P(.)$ is the predicted  probability of the actual label of the object. Other notations follow the conventions from above.
%For a image, $\rho^+_{\text{map}}$ boils down to the model's confidence score.
For an estimated saliency map, the value of $\rho^+_{\text{map}}$ improves if higher importance is attached to a smaller number of pixels that retain higher confidence of the model on the original label. In the extreme case of all the pixels deemed maximally important (saliency value 1), the score depicts the model's confidence on the object label. On the other hand, the value of $\rho^-_{\text{map}}$ decreases if lesser importance is attached by the saliency method to more pixels that do not influence the prediction confidence on the original label. Lower value of this metric is more desired.   

Combined $\rho^+_{\text{map}}$ and $\rho^-_{\text{map}}$ provide a comprehensive quantification of the quality of the saliency map. We provide results of our technique, Grad-CAM~\cite{selvaraju2017grad} and the recent Norm-Grad~\cite{rebuffi2020there} on our metrics in Table~\ref{tableScores}. {\color{black} Due to page limits, we provide further discussion on the proposed metrics in the supplementary material.}

%Particularly high values of our method for $\rho^+_{\text{map}}$ indicate accurate attribution of high importance to salient pixels. 

% \textcolor{red}{We have normalize the area (denominator of metric equations) by total area to keep the scores in appealing numerical range}

% We define two metrics to quantify the effectiveness of saliency maps in highlighting the important regions. 

% e quantify the effectiveness of saliency map by 

% \begin{align}
% 	\text{hit}_{density}^{\ell} = \frac{\mathrm{\text P} \Big(  \mathcal K({\bf I} \odot  \boldsymbol{S^{\ell}}) \Big)} 
% 	{ \sum_{j}^{} \sum_{k}^{} \boldsymbol{S_{j,k}^{\ell}}}  ~~s.t.~~
% 	 \mathcal K (\boldsymbol{I}) \rightarrow \ell
% 	\label{eq:score1}
% \end{align}

% \begin{align}
% 	\text{miss}_{density}^{\ell} = \frac{\mathrm{\text P} \Big(  \mathcal K({\bf I} \odot  ( \boldsymbol 1 - \boldsymbol{S^{\ell}})) \Big)} 
% 	{ \sum_{j}^{} \sum_{k}^{} (\boldsymbol 1 - \boldsymbol{S_{j,k}^{\ell}})}  ~~s.t.~~
% 	 \mathcal K (\boldsymbol{I}) \rightarrow \ell
% 	\label{eq:score2}
% \end{align}

%\subsection{Association }
\vspace{-2mm}
\section{CAMERAS for analysis}
\label{sec:CAMAnalysis}
\vspace{-1mm}
The precise CAMERAS results allow model analysis with backpropagation saliency in unprecedented details. Below, we present a few interesting examples.

\begin{figure}[t]
    \centering
    \includegraphics[width=0.45\textwidth]{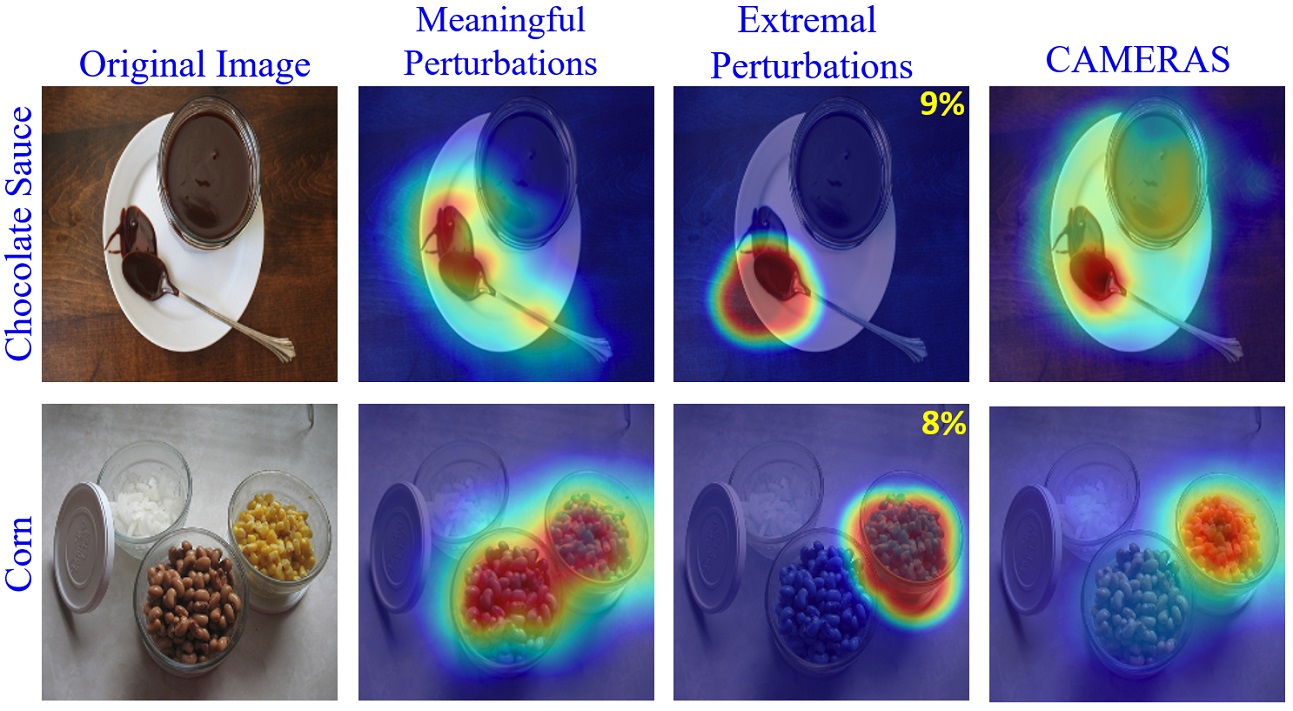}
    \caption{Proposed CAMERAS allows verification of label attribution even more precisely than the input  perturbation-based methods.  Meaningful \cite{fong2017interpretable} and Extremal \cite{fong2019understanding} perturbation methods require image-specific fine tuning of parameters. The latter is confined to 8\% and 9\% of the image area to achieve the results.}
    \label{fig:associationResults}
    \vspace{-3mm}
\end{figure}

\vspace{0.5mm}
\noindent{\bf The label attribution problem:} It is known  that deep visual classifiers sometimes learn incorrect association of labels with the objects in input images~\cite{fong2017interpretable}, \cite{fong2019understanding}. For instance, for the image of Chocolate Sauce in Fig.~\ref{fig:associationResults}, Inception is found to associate the said label to  spoon instead of the sauce in the cup~\cite{fong2017interpretable}, \cite{fong2019understanding}. This revelation was  possible only through the input perturbation-based methods for attribution due to their precise nature, however, only  after fine-tuning a list of parameters for the specific images. 
%E.g.~the perturbation is empirically confined to 9\% of the image area in Extremal perturbation~\cite{fong2019understanding} to restrict it to spoon.
CAMERAS is the first backpropagation saliency method to achieve this result, without requiring any image-specific fine tuning. Our method verifies the original results of the perturbation-based methods with an even better precision.

\begin{figure}[t]
    \centering
    \includegraphics[width=0.41\textwidth]{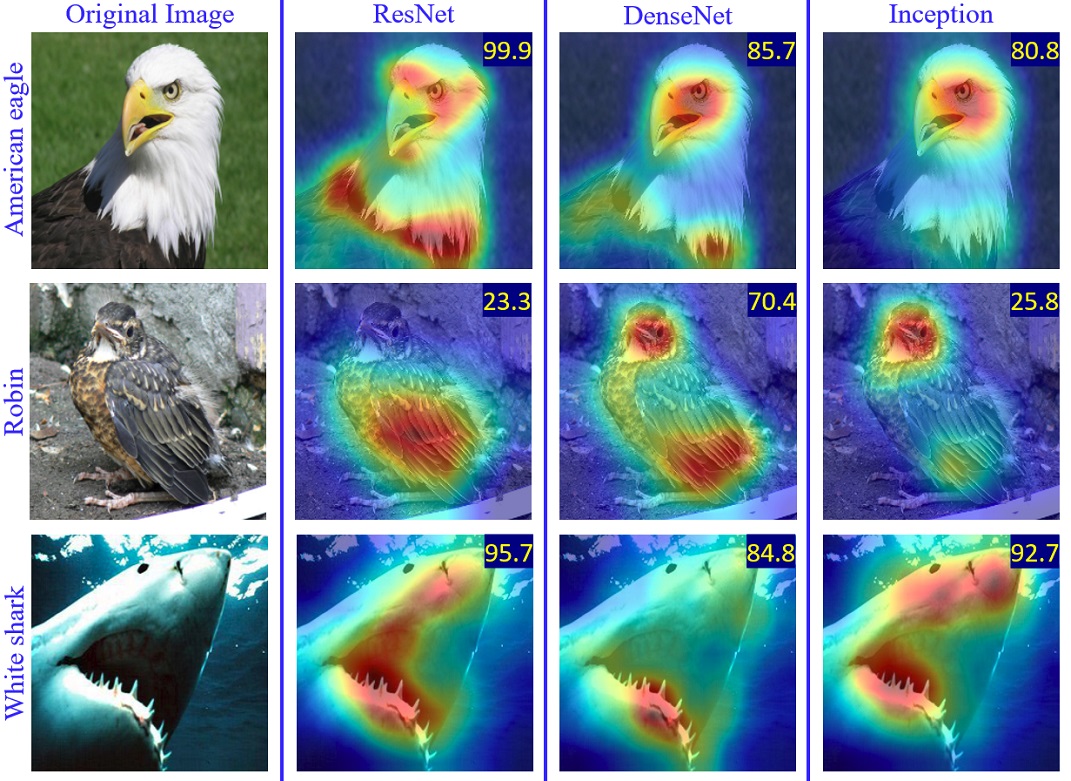}
    \caption{Precise saliency of CAMERAS reveals similarity in the level of attention on  fine-grained features causes similarity in the prediction confidence (given as percentages) of different models.}
    \label{fig:networkRelations}
    \vspace{-4mm}
\end{figure}

\vspace{0.5mm}
\noindent{\bf Prediction confidence:} In Fig.~\ref{fig:networkRelations}, CAMERAS results reveal that prediction confidence on individual images is often strongly influenced by a model's attention on fine-grained features. Different visual models may pay similar attention to the same features to achieve similar confidence scores. 

\begin{figure*}[t!]
    \centering
    \includegraphics[width=0.85\textwidth]{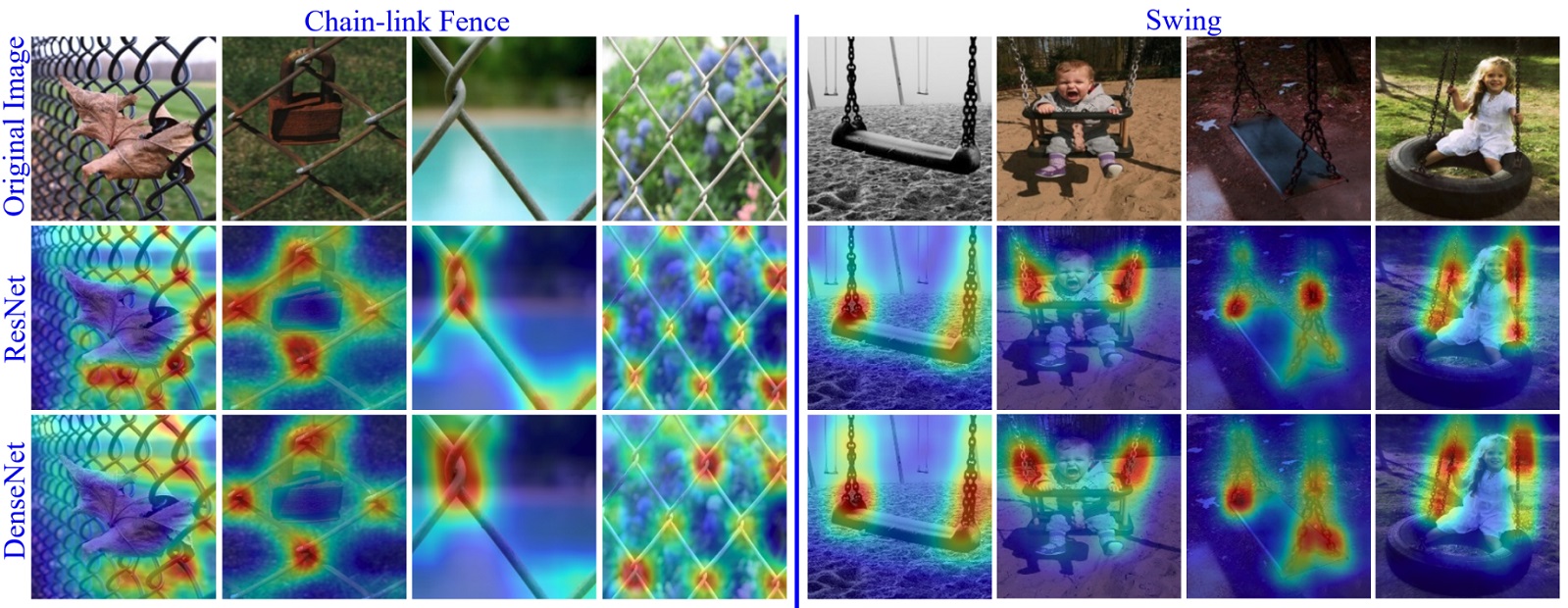}
    \caption{CAMERAS saliency maps reveal the differences of features learned by models to distinguish visually similar objects. }
    \label{fig:classRelations}
    \vspace{-3mm}
\end{figure*}

\vspace{0.5mm}
\noindent{\bf Discrimination of similar objects:}
Precise saliency mapping of CAMERAS also reveals clear  differences of the features learned by the models for visually similar objects. In Fig.~\ref{fig:classRelations}, we show the saliency maps for multiple examples of `Chain-link Fence' and `Swing'  for two high performing models. Notice how the models pay high  attention to the individual chain knots (left) as compared to the larger chain structures (right) to distinguish the two classes. These results also reinforce the importance of `not enforcing' any priors on the map (e.g.~smoothness~\cite{fong2019understanding}). The shown results provide the first instance of clear saliency differences between similar object features under backpropagation saliency mapping without external priors.

\vspace{-2mm}
\section{Adversarial Attack Enhancement}
\begin{figure}[thb!]
    \centering
    \includegraphics[width=0.47\textwidth]{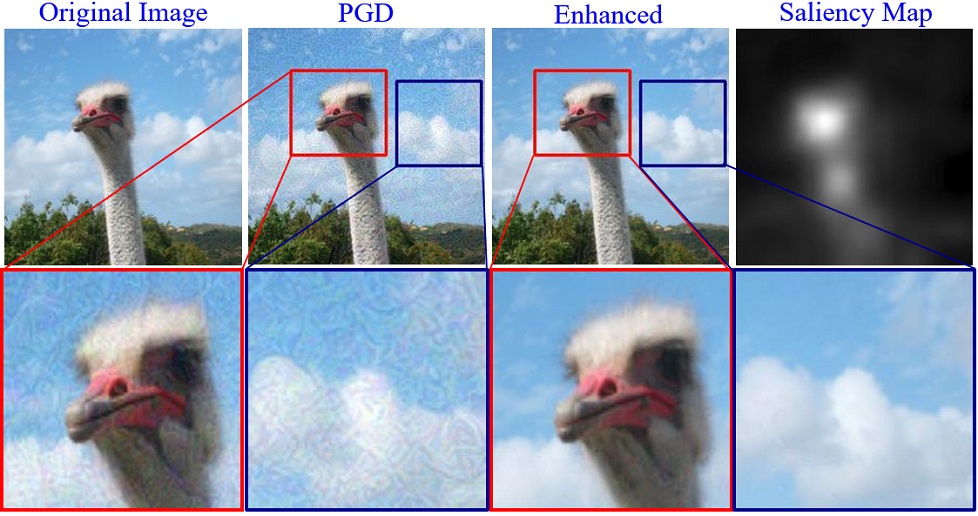}
    \caption{Enhancement of PGD with CAMERAS saliency. Confining the perturbation to high importance regions discovered by CAMERAS drastically reduces visual perceptibility of the attack while maintaining the  fooling ratio and confidence on the wrong labels. For clarity, $\ell_{\infty}$-norm of $12/255$ is chosen for perturbation.}
    \label{fig:attack}
    \vspace{-3mm}
\end{figure}

\vspace{-1mm}
Similar to the backpropagation saliency methods, most of the adversarial attacks on deep visual classifiers~\cite{akhtar2018threat} treat the models as differentiable programs. 
Using model gradients, they engineer additive noise (i.e.~perturbations) that alter model predictions on an input. To avoid attack suspicion, the perturbations must be kept norm-bounded. 
Projected Gradient Descent (PGD)~\cite{madry2017towards} is considered one of the strongest attacks~\cite{akhtar2018threat} that computes holistic perturbations to fool the models. Using PGD as an example, we show that precision saliency of CAMERAS can significantly enhance these attacks by confining the perturbations to the regions considered more salient by our method, see Fig.~\ref{fig:attack}.    

We iteratively solve for the following using the PGD
\begin{align}
	\min_{\boldsymbol p} \left(\mathcal J(\boldsymbol{I_p}, \ell_{ll}) + \beta~ ||  \boldsymbol{p} \odot (\boldsymbol 1 - \boldsymbol \Psi) ||_2\right), 
	 \label{eq:maskedAdversarialAttack}
\end{align}

where $\boldsymbol{I_p}$ is the perturbed image, $\mathcal{J}(.)$ is the cross entropy loss, $\ell_{ll}$ is the least likely label of the clean image, $\boldsymbol p$ is the perturbation, and $\beta = 50$  is an empirically chosen  scaling factor. In (\ref{eq:maskedAdversarialAttack}), we allow the perturbation signal to grow freely for our salient regions while restricting it in the other regions. By focusing only on the most important regions, we are able to drastically reduce the required perturbation norm. Maintaining 99.99\% fooling confidence for ResNet-50 on all images of ImageNet validation set, we successfully reduced the PGD perturbation norm by 56.5\% on average with our CAMERAS enhancement. {\color{black} See the supplementary material for more details.} Other adversarial attacks can also be enhanced similarly with CAMERAS.   

% \vspace{-1mm}
\subsection{Sanity test with adversarial perturbation}
\label{sec:Pbst}
\vspace{-1.5mm}
Gradient-based adversarial attacks algorithmically compute  minimal perturbations to image pixels to maximally change the model prediction. This objective coincides with the  objective of image saliency computation, thereby providing a natural sanity check for the saliency methods.  That is, the effects of corruption  with an adversarial perturbation to image pixels should correspond to the importance of the pixels  identified by the saliency method. Leveraging this fact, we develop a sanity check for saliency methods that operates on pixel-by-pixel basis, which  is more suited for precise image saliency. We provide details of the test in the supplementary material due to page limits.

\vspace{-2mm}
\section{Conclusion}
\label{sec:Conc}
\vspace{-1mm}
We introduced CAMERAS to compute precise saliency maps using the gradient backpropagation strategy. Our technique is shown to preserve the sanity of the computed saliency maps by avoiding external influence and priors over the maps. High precision of our saliency maps allow better  explanation of deep visual model predictions. We also demonstrated application of CAMERAS to enhance adversarial attacks, and used this to introduce a new sanity check for high-fidelity saliency methods.   

\vspace{2mm}
\textbf{Acknowledgment} This  research  was  supported  by ARC Discovery Grant DP190102443 and partially by DP150100294 and DP150104251. The Titan V used in our experiments was donated by NVIDIA corporation.

\balance
\newpage
{\small
\bibliographystyle{ieee_fullname}
\bibliography{egpaper_for_review}

\begin{thebibliography}{10}\itemsep=-1pt

\bibitem{adadi2018peeking}
Amina Adadi and Mohammed Berrada.
\newblock Peeking inside the black-box: A survey on explainable artificial
  intelligence (xai).
\newblock {\em IEEE Access}, 6:52138--52160, 2018.

\bibitem{adebayo2018sanity}
Julius Adebayo, Justin Gilmer, Michael Muelly, Ian Goodfellow, Moritz Hardt,
  and Been Kim.
\newblock Sanity checks for saliency maps.
\newblock In {\em Advances in Neural Information Processing Systems}, pages
  9505--9515, 2018.

\bibitem{akhtar2018threat}
Naveed Akhtar and Ajmal Mian.
\newblock Threat of adversarial attacks on deep learning in computer vision: A
  survey.
\newblock {\em IEEE Access}, 6:14410--14430, 2018.

\bibitem{bach2015pixel}
Sebastian Bach, Alexander Binder, Gr{\'e}goire Montavon, Frederick Klauschen,
  Klaus-Robert M{\"u}ller, and Wojciech Samek.
\newblock On pixel-wise explanations for non-linear classifier decisions by
  layer-wise relevance propagation.
\newblock {\em PloS one}, 10(7):e0130140, 2015.

\bibitem{chen2018encoder}
Liang-Chieh Chen, Yukun Zhu, George Papandreou, Florian Schroff, and Hartwig
  Adam.
\newblock Encoder-decoder with atrous separable convolution for semantic image
  segmentation.
\newblock In {\em Proceedings of the European conference on computer vision
  (ECCV)}, pages 801--818, 2018.

\bibitem{dabkowski2017real}
Piotr Dabkowski and Yarin Gal.
\newblock Real time image saliency for black box classifiers.
\newblock In {\em Advances in Neural Information Processing Systems}, pages
  6967--6976, 2017.

\bibitem{pascal-voc-2007}
M. Everingham, L. Van~Gool, C.~K.~I. Williams, J. Winn, and A. Zisserman.
\newblock The {PASCAL} {V}isual {O}bject {C}lasses {C}hallenge 2007 {(VOC2007)}
  {R}esults.
\newblock
  http://www.pascal-network.org/challenges/VOC/voc2007/workshop/index.html.

\bibitem{fong2019understanding}
Ruth Fong, Mandela Patrick, and Andrea Vedaldi.
\newblock Understanding deep networks via extremal perturbations and smooth
  masks.
\newblock In {\em Proceedings of the IEEE International Conference on Computer
  Vision}, pages 2950--2958, 2019.

\bibitem{fong2017interpretable}
Ruth~C Fong and Andrea Vedaldi.
\newblock Interpretable explanations of black boxes by meaningful perturbation.
\newblock In {\em Proceedings of the IEEE International Conference on Computer
  Vision}, pages 3429--3437, 2017.

\bibitem{he2016deep}
Kaiming He, Xiangyu Zhang, Shaoqing Ren, and Jian Sun.
\newblock Deep residual learning for image recognition.
\newblock In {\em Proceedings of the IEEE conference on computer vision and
  pattern recognition}, pages 770--778, 2016.

\bibitem{jalwana2020attack}
Mohammad~AAK Jalwana, Naveed Akhtar, Mohammed Bennamoun, and Ajmal Mian.
\newblock Attack to explain deep representation.
\newblock In {\em Proceedings of the IEEE/CVF Conference on Computer Vision and
  Pattern Recognition}, pages 9543--9552, 2020.

\bibitem{kapishnikov2019xrai}
Andrei Kapishnikov, Tolga Bolukbasi, Fernanda Vi{\'e}gas, and Michael Terry.
\newblock Xrai: Better attributions through regions.
\newblock In {\em Proceedings of the IEEE International Conference on Computer
  Vision}, pages 4948--4957, 2019.

\bibitem{kindermans2019reliability}
Pieter-Jan Kindermans, Sara Hooker, Julius Adebayo, Maximilian Alber, Kristof~T
  Sch{\"u}tt, Sven D{\"a}hne, Dumitru Erhan, and Been Kim.
\newblock The (un) reliability of saliency methods.
\newblock In {\em Explainable AI: Interpreting, Explaining and Visualizing Deep
  Learning}, pages 267--280. Springer, 2019.

\bibitem{kindermans2016investigating}
Pieter-Jan Kindermans, Kristof Sch{\"u}tt, Klaus-Robert M{\"u}ller, and Sven
  D{\"a}hne.
\newblock Investigating the influence of noise and distractors on the
  interpretation of neural networks.
\newblock {\em arXiv preprint arXiv:1611.07270}, 2016.

\bibitem{krizhevsky2012imagenet}
Alex Krizhevsky, Ilya Sutskever, and Geoffrey~E Hinton.
\newblock Imagenet classification with deep convolutional neural networks.
\newblock In {\em Advances in neural information processing systems}, pages
  1097--1105, 2012.

\bibitem{lin2014microsoft}
Tsung-Yi Lin, Michael Maire, Serge Belongie, James Hays, Pietro Perona, Deva
  Ramanan, Piotr Doll{\'a}r, and C~Lawrence Zitnick.
\newblock Microsoft coco: Common objects in context.
\newblock In {\em European conference on computer vision}, pages 740--755.
  Springer, 2014.

\bibitem{long2015fully}
Jonathan Long, Evan Shelhamer, and Trevor Darrell.
\newblock Fully convolutional networks for semantic segmentation.
\newblock In {\em Proceedings of the IEEE conference on computer vision and
  pattern recognition}, pages 3431--3440, 2015.

\bibitem{madry2017towards}
Aleksander Madry, Aleksandar Makelov, Ludwig Schmidt, Dimitris Tsipras, and
  Adrian Vladu.
\newblock Towards deep learning models resistant to adversarial attacks.
\newblock {\em arXiv preprint arXiv:1706.06083}, 2017.

\bibitem{mahendran2016salient}
Aravindh Mahendran and Andrea Vedaldi.
\newblock Salient deconvolutional networks.
\newblock In {\em European Conference on Computer Vision}, pages 120--135.
  Springer, 2016.

\bibitem{petsiuk2018rise}
Vitali Petsiuk, Abir Das, and Kate Saenko.
\newblock Rise: Randomized input sampling for explanation of black-box models.
\newblock {\em arXiv preprint arXiv:1806.07421}, 2018.

\bibitem{rebuffi2020there}
Sylvestre-Alvise Rebuffi, Ruth Fong, Xu Ji, and Andrea Vedaldi.
\newblock There and back again: Revisiting backpropagation saliency methods.
\newblock In {\em Proceedings of the IEEE/CVF Conference on Computer Vision and
  Pattern Recognition}, pages 8839--8848, 2020.

\bibitem{redmon2017yolo9000}
Joseph Redmon and Ali Farhadi.
\newblock Yolo9000: better, faster, stronger.
\newblock In {\em Proceedings of the IEEE conference on computer vision and
  pattern recognition}, pages 7263--7271, 2017.

\bibitem{ren2015faster}
Shaoqing Ren, Kaiming He, Ross Girshick, and Jian Sun.
\newblock Faster r-cnn: Towards real-time object detection with region proposal
  networks.
\newblock In {\em Advances in neural information processing systems}, pages
  91--99, 2015.

\bibitem{ribeiro2016should}
Marco~Tulio Ribeiro, Sameer Singh, and Carlos Guestrin.
\newblock " why should i trust you?" explaining the predictions of any
  classifier.
\newblock In {\em Proceedings of the 22nd ACM SIGKDD international conference
  on knowledge discovery and data mining}, pages 1135--1144, 2016.

\bibitem{selvaraju2017grad}
Ramprasaath~R Selvaraju, Michael Cogswell, Abhishek Das, Ramakrishna Vedantam,
  Devi Parikh, and Dhruv Batra.
\newblock Grad-cam: Visual explanations from deep networks via gradient-based
  localization.
\newblock In {\em Proceedings of the IEEE International Conference on Computer
  Vision}, pages 618--626, 2017.

\bibitem{shrikumar2017learning}
Avanti Shrikumar, Peyton Greenside, and Anshul Kundaje.
\newblock Learning important features through propagating activation
  differences.
\newblock {\em arXiv preprint arXiv:1704.02685}, 2017.

\bibitem{simonyan2014deep}
Karen Simonyan, Andrea Vedaldi, and Andrew Zisserman.
\newblock Deep inside convolutional networks: Visualising image classification
  models and saliency maps.
\newblock 2014.

\bibitem{simonyan2014very}
Karen Simonyan and Andrew Zisserman.
\newblock Very deep convolutional networks for large-scale image recognition.
\newblock {\em arXiv preprint arXiv:1409.1556}, 2014.

\bibitem{smilkov2017smoothgrad}
Daniel Smilkov, Nikhil Thorat, Been Kim, Fernanda Vi{\'e}gas, and Martin
  Wattenberg.
\newblock Smoothgrad: removing noise by adding noise.
\newblock {\em arXiv preprint arXiv:1706.03825}, 2017.

\bibitem{springenberg2014striving}
Jost~Tobias Springenberg, Alexey Dosovitskiy, Thomas Brox, and Martin
  Riedmiller.
\newblock Striving for simplicity: The all convolutional net.
\newblock {\em arXiv preprint arXiv:1412.6806}, 2014.

\bibitem{sundararajan2017axiomatic}
Mukund Sundararajan, Ankur Taly, and Qiqi Yan.
\newblock Axiomatic attribution for deep networks.
\newblock {\em arXiv preprint arXiv:1703.01365}, 2017.

\bibitem{tsipras2018robustness}
Dimitris Tsipras, Shibani Santurkar, Logan Engstrom, Alexander Turner, and
  Aleksander Madry.
\newblock Robustness may be at odds with accuracy.
\newblock {\em arXiv preprint arXiv:1805.12152}, 2018.

\bibitem{zeiler2014visualizing}
Matthew~D Zeiler and Rob Fergus.
\newblock Visualizing and understanding convolutional networks.
\newblock In {\em European conference on computer vision}, pages 818--833.
  Springer, 2014.

\bibitem{zhang2018top}
Jianming Zhang, Sarah~Adel Bargal, Zhe Lin, Jonathan Brandt, Xiaohui Shen, and
  Stan Sclaroff.
\newblock Top-down neural attention by excitation backprop.
\newblock {\em International Journal of Computer Vision}, 126(10):1084--1102,
  2018.

\bibitem{zhou2016learning}
Bolei Zhou, Aditya Khosla, Agata Lapedriza, Aude Oliva, and Antonio Torralba.
\newblock Learning deep features for discriminative localization.
\newblock In {\em Proceedings of the IEEE conference on computer vision and
  pattern recognition}, pages 2921--2929, 2016.

\bibitem{zintgraf2017visualizing}
Luisa~M Zintgraf, Taco~S Cohen, Tameem Adel, and Max Welling.
\newblock Visualizing deep neural network decisions: Prediction difference
  analysis.
\newblock {\em arXiv preprint arXiv:1702.04595}, 2017.

\end{thebibliography}
}

\end{document}